\documentclass[twoside,11pt]{article}

\usepackage{jmlr2e}

\usepackage{epsf}
\usepackage{times}
\usepackage{epsfig}
\usepackage{graphicx}
\usepackage{amsmath}
\usepackage{amssymb}
\usepackage{algorithm}
\usepackage{algorithmic}
\usepackage{multirow}
\usepackage[section] {placeins} 
\usepackage{color}

\jmlrheading{}{2013}{1-37}{03/08}{}{Hamed Masnadi-Shirazi }

\ShortHeadings{Refinement revisited }{Masnadi-Shirazi }
\firstpageno{1}

\begin{document}

\title{Refinement revisited with connections to Bayes error, conditional entropy and calibrated classifiers}

\author{\name Hamed Masnadi-Shirazi \email hmasnadi@shirazu.ac.ir \\
       \addr School of Electrical and Computer Engineering, \\
				Shiraz University, \\
				Shiraz, Iran }
				
\editor{Lawrence Saul}

\maketitle
%
%
%
%
%
%
%
%

\begin{abstract}
The concept of refinement from probability 
elicitation   is considered for proper scoring rules. Taking directions from the axioms of probability,
refinement is further clarified using a Hilbert space interpretation 
and reformulated into the underlying data
distribution setting where connections to maximal marginal diversity and
conditional entropy are considered and used to derive measures that provide
arbitrarily tight bounds on the Bayes error.  Refinement is also reformulated
into the classifier output setting and its connections to calibrated classifiers
and proper margin losses are established.
\end{abstract}



\section{Introduction}
The concept of refinement can be traced back to a well known partition of the Brier (or quadratic) score in early works by \cite{Murphy1972} but was explicitly defined and generalized for all proper scoring rules in a series of  seminal papers by DeGroot and Fienberg \cite{DeGrootBook,DeGroot}. This concept is also well known under different names depending on the literature. In the forecasting and meteorology literature it is know as sharpness \cite{Sanders1963,Tilmann2007} or  
resolution \cite{Brocker2009} and in the probability elicitation  literature \cite{Savage} it is also know as sufficiency \cite{DeGroot,DeGrootBook,Schervish1989}. 
This concept has also been studied  most recently in the meteorology and forecasting literature in papers such as \cite{Tilmann2005,Wilks2006,Tilmann2007,Brocker2009}.

Despite the fact that refinement is closely related to proper scoring rules and  calibrated loss functions it has remained largely restricted to
the probability elicitation and forecasting literature. In this paper we initially briefly review the concepts of calibration and refinement. The concept of refinement will be emphasized and explained using the original works of DeGroot and Fienberg \cite{DeGrootBook}. We will then proceed to bring three different yet closely interlocked arguments that will each initially seem to refute the validity of the refinement concept, but will instead after a subtle clarification,  lead to the generalization of the refinement concept and establish its  connections to  Bayes error, maximum margin diversity and conditional entropy in feature selection \cite{NUNOMaxDiversityNIPS,FastBinaryFeatureSelection, Pengfeatureselection, NunoNaturalFeatures},  and classification with Bayes calibrated loss functions \cite{friedman,Zhang,Buja,HamedNunoLossDesign,Reid} among others. Specifically, the original refinement definition on the probability elicitation setting will be extended to the classifier output setting and underlying data distribution setting. 

A series of results are presented by extending refinement to the underlying data distribution setting  which show  that conditional entropy and maximum margin diversity used in feature selection are a special case of refinement using the logistic score function. A number of other novel refinement measures based on other score functions are derived along with  \emph{conditional refinement} which can be used for feature selection and ranking. Refinement is also related to the Bayes error. A number of well known bounds on the Bayes error such as the  Battacharyy bound \cite{book:Fukunaga}  the asymptotic nearest neighbor bound \cite{book:Fukunaga,NNClassification} and the  Jensen–Shannon divergence \cite{JenShannonLin} are shown to be special cases of refinement measures. Other novel bounds on the Bayes error are derived using the refinement interpretation along with a method for deriving arbitrarily tight bounds on the Bayes error.

Extending refinement to the classifier output setting allows for a statistically rigorous parallel to the classifier margin which we call \emph{classifier marginal density} which allows for the ranking of calibrated classifiers simply based on their outputs. We also show how each  calibrated loss function has a corresponding 
refinement measure and derive a number of such novel measures.

Refinement is also further studied in its original probability elicitation setting and a Hilbert space and inner product interpretation is provided. 
The inner product interpretation leads to further insight into refinement using different symmetric scoring rules.

The paper is organized as follows. In Section-\ref{sec:RefElic} a review of the refinement concept in probability elicitation is provided. In Section-\ref{sec:FurtherRefConcpt} the refinement concept is further analyzed from the perspective of the axioms of probability which leads to a novel refinement formulation in the underlying data distribution setting and connections to the Bayes error. In Section-\ref{sec:RefCondEntropy} refinement and its connections to maximal marginal diversity and conditional entropy are considered.
Connections to calibrated classifiers are considered in Section-\ref{sec:RefCaliClass} . 
In Sections- \ref{sec:FurtherOrigElic},\ref{sec:FurtherClassOut}, and \ref{sec:FurtherUnderlyData} refinement is further studied in its original setting,
the proposed classifier output and underlying data distribution settings, respectively. Finally, in Section-\ref{sec:TighterBounds} refinement in the underlying data distribution setting is used to derive measures that provide arbitrarily tighter bounds on the Bayes error. Summary and conclusions are provided in Section-\ref{sec:conclusion}.


\section{Refinement In Probability Elicitation}
\label{sec:RefElic}

In probability elicitation \cite{Savage} a forecaster produces a  probability estimate  $\hat \eta$ of the occurrence of event $y=1$ where $y \in \{1,~-1\}$, such as a
weatherman predicting that it will rain ($y=1$)
tomorrow. $\eta=P(1|\hat \eta)$ is the actual relative frequency of event $y=1$ (rain) among
those days which the forecaster's prediction was $\hat \eta$.  A forecaster is said to be
 \textit{calibrated} if $\eta=\hat \eta$ for all $\hat \eta$, meaning that the weatherman is skilled and trustworthy. In other words it actually rains
$\eta=\hat \eta$ percent of the time when he predicts the chance of rain is $\hat \eta$.

It has been argued in \cite{DeGroot2,DeGrootBook,Dawid1981} that a calibrated forecaster is not necessarily a good forecaster or an informative and useful one and that another concept called \emph{refinement} is also needed to evaluate forecasters.
Intuitively, let $s(\hat \eta)$ denote the probability density function of the forecaster's predictions, then the more concentrated the probability density function $s(\hat \eta)$ is around the values $\hat \eta=0$ and $\hat \eta=1$ the more \textit{refined} the forecaster is. To further demonstrate the concept of refinement, it is useful to consider the following slightly modified example taken from \cite{DeGrootBook}. Consider two calibrated weather forecasters $A$ and $B$ working at a location where the expected probability of rain is $\mu=0.5$ on any given day. Weatherman $A$ is such that
\begin{eqnarray}
&&s_{A}(\mu)=1 \\
&&s_{A}(\hat \eta)=0 \hspace{0.2in} \mbox{for} \hspace{0.2in} \hat \eta \ne \mu
\end{eqnarray}
and weatherman $B$ is such that
\begin{eqnarray}
&&s_{B}(1)=\mu \\
&&s_{B}(0)=1-\mu \\
&&s_{B}(\hat \eta)=0 \hspace{0.2in} \mbox{for} \hspace{0.2in} \hat \eta \ne 0,1.
\end{eqnarray}
Both forecasters can be calibrated. To demonstrate this, assume that both weathermen make $100$ predictions.  Weatherman A predicts that the chance of rain is $\hat \eta=0.5$ all the time. If it actually rains as expected on $50$ days  we have $\eta=\frac{50}{100}=0.5$ so $\hat \eta =\eta$ and A is calibrated. In the case of weatherman B the predictions are 1) chance of rain is $\hat \eta=1$ on $50$ days and 2) chance of rain is $\hat \eta=0$ on the other $50$ days. If it actually rains on the $50$ days B predicted rain  then $\eta=\frac{50}{50}=1$ and if it actually does not rain when B predicted no rain then $\eta=\frac{0}{50}=0$. 
In either case $\hat \eta =\eta$ and B is also calibrated. 
  
Although we have shown that both A and B are calibrated forecasters, it is acceptable to say that the forecasts made by A are useless  while forecaster B is the
ideal weatherman in the sense that he only makes definite predictions of chance of rain is $0$  or chance of rain is $1$
and is always correct. On the other hand, forecaster A always makes the conservative but useless prediction that chance of rain is $0.5$. We say that weatherman A is the least-refined forecaster and that weatherman B is the most-refined forecaster \cite{DeGrootBook}.
This leads to the argument that well calibrated forecasters can be compared based on  their refinement \cite{DeGroot}.

Before providing a formal measure of refinement, proper scoring functions need to be introduced. 
A scoring function is such that a score of $I_1(\hat \eta)$ is
attained if the forecaster predicts $\hat \eta$ and event $y=1$ actually happens and a score of $I_{-1}(\hat \eta)$ is attained if event $y=-1$  happens. 
$I_1(\hat \eta)$  is an increasing functions of $\hat \eta$ and $I_{-1}(\hat \eta)$ is decreasing in $\hat \eta$. Since the relative frequency with which the forecaster makes the prediction $\hat \eta$ is $s(\hat \eta)$, the 
expected score of the forecaster over all $\hat \eta$ and $y$ is
\begin{eqnarray}
\int_{\hat \eta} s(\hat \eta)[\eta I_1(\hat \eta)+(1-\eta)I_{-1}(\hat \eta)] d(\hat \eta),
\end{eqnarray}
and the expected score for a given $\hat \eta$ is
\begin{eqnarray}
\label{eq:CondExpectScore}
I(\eta,\hat \eta)=\eta I_1(\hat \eta)+(1-\eta)I_{-1}(\hat \eta).
\end{eqnarray}

The score function is denoted as  \emph{strictly proper}
if $I_1(\hat \eta)$ and $I_{-1}(\hat \eta)$ are such
that the  expected score of (\ref{eq:CondExpectScore}) is maximized when $\hat \eta=\eta$ or in other words  
\begin{eqnarray}
\label{eq:DefJ}
I(\eta,\hat \eta) \le I(\eta,\eta)=J(\eta).
\end{eqnarray}

It can be shown \cite{Savage} that a score function is strictly proper  if and 
only if the \emph{maximal reward function} $J(\eta)$ is strictly convex and 
\begin{eqnarray}
  \label{eq:Is}
  I_1(\eta) &=& J(\eta) + (1-\eta) J^\prime(\eta) \label{eq:I1}  \label{eq:I1Jprime} \\
  I_{-1}(\eta) &=& J(\eta) -\eta J^\prime(\eta) \label{eq:I-1}. \label{eq:I2Jprime} 
\end{eqnarray}


A formal definition of refinement can be provided when considering the proper scoring function $I_y$. The expected score $S_{I_y}$ can  be written as
\begin{eqnarray}
S_{I_y}&&=\int_{\hat \eta} s(\hat \eta) \sum_y P(y|\hat \eta)I_{y}(\hat \eta) d(\hat \eta)  \\
&&=\int_{\hat \eta} s(\hat \eta) \Bigl( P(1|\hat \eta)I_1(\hat \eta) + P(-1|\hat \eta)I_{-1}(\hat \eta))\Bigr) d(\hat \eta). \nonumber
\end{eqnarray}
By simply adding and subtracting 
$s(\hat \eta)[P(1|\hat \eta)I_1(\eta) + P(-1|\hat \eta)I_{-1}(\eta)]$, we can
dissect any expected score $S_{I_y}$ of a forecaster
into two parts of $S_{Calibration}$ and $S_{Refinement}$ that are measures of calibration and refinement \cite{DeGroot}  
\begin{eqnarray}
&& S_{I_y}=\int_{\hat \eta} s(\hat \eta) \sum_y P(y|\hat \eta)I_{y}(\hat \eta) d(\hat \eta)  \\
&&=\int_{\hat \eta} s(\hat \eta) \Bigl( P(1|\hat \eta)I_1(\hat \eta) + P(-1|\hat \eta)I_{-1}(\hat \eta))\Bigr) d(\hat \eta) \nonumber \\ 
&&= \int_{\hat \eta} s(\hat \eta) \Bigl[ P(1|\hat \eta)\Bigl\{I_1(\hat \eta)-I_1(\eta)\Bigr\} +  
P(-1|\hat \eta)\Bigl\{I_{-1}(\hat \eta)-I_{-1}(\eta) \Bigr\} \Bigr] d(\hat \eta)  \nonumber  \\ 
&& + \int_{\hat \eta} s(\hat \eta)\Bigl[P(1|\hat \eta)I_1(\eta) + P(-1|\hat \eta)I_{-1}(\eta)\Bigr] d(\hat \eta) \nonumber \\
&&=S_{Calibration}+S_{Refinement}. \nonumber
\end{eqnarray}

Recall that $I(\hat \eta,\eta) \le I(\eta,\eta)=J(\eta)$  
so that $S_{Calibration}$ has a maximum equal to zero when the forecaster is calibrated ($\hat \eta=\eta$) and is negative otherwise.

The second term $S_{Refinement}$ can be simplified to 
\begin{eqnarray}
\label{eq:RefOrig}
S_{Refinement}\!\!\!\!\!\!&&=\int_{\hat \eta} s(\hat \eta)\Bigl[P(1|\hat \eta)I_1(\eta) + P(-1|\hat \eta)I_{-1}(\eta)\Bigr] d(\hat \eta) \\
&&=\int_{\hat \eta} s(\hat \eta)\Bigl[\eta I_1(\eta) + (1-\eta) I_{-1}(\eta)\Bigr] d(\hat \eta) \nonumber \\
&&=\int_{\hat \eta} s(\hat \eta)J(\eta) d(\hat \eta). \nonumber
\end{eqnarray}

Note that $J(\eta)=J(P(1|\hat \eta))$ is  a convex function of   $\hat \eta$ over the $[0~1]$ interval.
Intuitively, the more concentrated $s(\hat \eta)$ is near $0$ and $1$ the larger the $s(\hat \eta)J(\eta)$ term will become. In other words $S_{Refinement}$ will increase as $\hat \eta(\bf{x})$ becomes more refined \cite{DeGroot}.
We will formalize this and present the inner product interpretation of refinement in Section-\ref{sec:FurtherOrigElic}. 

As an example, the expected score of the strictly proper Brier score (BS) (or least squares) $I_{y'}=(\hat \eta - y')^2$ where $y'=\frac{y+1}{2}$,  can be expressed as a measure of calibration and refinement \cite{Murphy1972,DeGroot}
\begin{eqnarray}
&& S_{BS}=-\int_{\hat \eta} s(\hat \eta) \Bigl( P(1|\hat \eta)(\hat \eta -1)^2 + P(-1|\hat \eta)(\hat \eta)^2 \Bigr) d(\hat \eta)\\
&&= -\int_{\hat \eta} s(\hat \eta) \Bigl(\hat \eta - P(1|\hat \eta)\Bigr)^2 d(\hat \eta)+  
\int_{\hat \eta} s(\hat \eta)P(1|\hat \eta)\Bigl(P(1|\hat \eta)-1\Bigr) d(\hat \eta) \nonumber \\
&&=S_{Calibration}+S_{Refinement}. \nonumber
\end{eqnarray}
The expected score $S_{I_y}$  is maximized when the forecaster is calibrated $\hat \eta=P(1|\hat \eta)=\eta$ and  the distribution of predictions $s(\hat \eta)$ are mostly concentrated around $0$ and $1$ since $P(1|\hat \eta)\Bigl(P(1|\hat \eta)-1\Bigr)=\eta(\eta-1)$ is a symmetric convex function of $\eta$ on this interval with minimum at $\eta=\frac{1}{2}$ and maximums at $\eta=0$ and $\eta=1$ \cite{DeGroot}.

\section{ Further Analysis Of The Refinement Concept}
\label{sec:FurtherRefConcpt}
In this section we present a series of three arguments from different angles that further clarify and extend the concept of refinement.
The first is an argument based on Cox's theorem on subjective probability that basically points out a subtle yet important flaw in the assumptions that might be made in understanding the refinement concept.  

\subsection{Argument based on the basic desiderata of probability}
\label{sec:RefAxioms}
A  forecaster is simply producing subjective probabilities. It is well understood that subjective probabilities  are based on the axioms of Cox's theory which are  elegantly presented  as the desiderata of probability in the form of three logical statements in \cite{book:Jaynes}.  
It is the failure to strictly follow these requisites that has led to many unnecessary errors, paradoxes and controversies in probability. 
Here we show that the concept of refinement might seem to contradict the third desiderata of probability if not presented correctly, namely that of \emph{consistency}. This requires that 1) if a conclusion can be reasoned out in more than one way, then every possible way must lead to the same result, 2) the forecaster always takes into account all  the evidence it has relevant to the question and  does not arbitrarily ignore some of the information and 3) if in two problems the forecasters state of knowledge is the same, then it must assign the same probabilities in both. It is also important to note that subjective probability and their logic does not depend on the person or machine making them. Anyone who has the same information  but comes to a different probability assignment is necessarily violating one of the desiderata of probability \cite{book:Jaynes}.

Ignoring the above requisites can lead to a misunderstanding or contradiction when considering the concept of refinement. This can best be presented with an example similar to that  in Section-\ref{sec:RefElic}. Assume that two \emph{calibrated} forecasters A and B have access to the same information, for the sake of argument we assume this to be data $x$ in the form of air pressure readings which is a good indicator for predicting rain. Also assume that the actual probability of rain given  air pressure $x$ is known to be $P(1|x)=0.7$. In terms of forecasters, the \emph{consistency} property requires that each $x$ lead to a corresponding forecast $\hat \eta$ (and $\eta$) and that no $x$ lead to more than one forecast $\hat \eta$ (and $\eta$). In other words $\hat \eta$ and $\eta$ are  functions of the information $x$ such that we can write  $\hat \eta(x)$ and $\eta(x)$.
 
Let Forecaster A make the prediction that chance of rain is $\hat \eta_A(x)=1$ and forecaster B make the prediction that chance of rain is $\hat \eta_B(x)=0.7$.  It might initially seem that forecaster A is more refined than forecaster B, but in fact the consistency principle of probability elicitation is being violated. In other words, since both forecasters  are basing their forecasts on the same information $x$, they should both make identical predictions. 


We extend the concept of forecasters and require two more reasonable properties from a forecaster. First, a forecaster should be responsive. In other words different information must lead to a different forecast. Formally, we require that if the information $x_1 \ne x_2$ then $\hat \eta(x_1) \ne \hat \eta(x_2)$ and $\eta(x_1) \ne \eta(x_2)$ .  This is equivalent by definition to requiring that $\hat \eta(x)$ and $\eta(x)$ be one-to-one functions. Second, a forecaster should be encompassing and any forecast should be possible. Formally, their exists a corresponding $x$ for any $\hat \eta$ and $\eta$. This is equivalent by definition to requiring that $\hat \eta(x)$ and $\eta(x)$ be onto functions. Both required properties can be summarized by equivalently requiring that $\hat \eta(x)$ and $\eta(x)$ be invertible functions. The immediate consequence of invertibility is that 
\begin{eqnarray}
\label{eq:invrtPeta}
\eta(x)=P(1|\hat \eta(x))=P(1|x).
\end{eqnarray}

This in turn leads to another contradiction in the example above  meaning that forecaster A is not actually calibrated. If as stated $P(1|x)=0.7$ then $\eta(x)=P(1|\hat \eta(x))=P(1|x)=0.7$ while  forecaster A predicted $\hat \eta_A(x)=1 \ne \eta(x)$ i.e. forecaster A is not calibrated as initially claimed.
Forecaster B, on the other hand, is verifiably calibrated. 

\subsection{Extending refinement to the underlying data distribution setting}  
\label{sec:ExtendRefToUnderlyData}
The discussion and example presented in Section \ref{sec:RefAxioms} suggest that a forecaster and its measure of refinement depend on the underlying data distribution
$P(1|x)$ from which the forecasts are established. This can be formally presented by writing the expected score as
\begin{eqnarray}
&&E_{\hat \eta,Y}[I_y(\hat \eta)]=  \int_{\hat \eta} s(\hat \eta) \sum_y P(y|\hat \eta)I_y(\hat \eta) d(\hat \eta) \\
&& =\int_{\hat \eta} s(\hat \eta) \sum_y \frac{P(\hat \eta|y)P(y)}{s(\hat \eta)} I_y(\hat \eta) d(\hat \eta) \nonumber\\
&&  =\int_{\hat \eta} \Bigl[ P(\hat \eta|1)P(1)I_1(\hat \eta)+P(\hat \eta|-1)P(-1)I_{-1}(\hat \eta)\Bigr] d(\hat \eta) \nonumber\\
&&  =\int_{X} \Bigl[ \frac{P(x|1)}{\hat \eta'}P(1)I_1(\hat \eta)+\frac{P(x|-1)}{\hat \eta'}P(-1)I_{-1}(\hat \eta) \Bigr](\hat \eta'dx))  \nonumber\\
&&  =\int_{X} \Bigl[ P(x|1)P(1)I_1(\hat \eta)+P(x|-1)P(-1)I_{-1}(\hat \eta)\Bigr] dx  \nonumber\\
&&  =\int_{X} P_X(x) \sum_y P(y|x)I_y(\hat \eta) dx \nonumber
\end{eqnarray} 
where $\hat \eta'=\hat \eta'(x)=\frac{d \hat \eta(x)}{dx}$  
and we have made use of the change of variable theory from calculus and function of random variable theory from probability theory. Using this theory demands that $\hat \eta(x)$ be an invertible function as  previously required for a forecaster.
 
The refinement term (\ref{eq:RefOrig}) can also be similarly reduced to
\begin{eqnarray}
\label{eq:RefDataSetting}
S_{Refinement}\!\!\!\!\!\!&&=\int_{\hat \eta} s(\hat \eta)\Bigl[P(1|\hat \eta)I_1(\eta) + P(-1|\hat \eta)I_{-1}(\eta)\Bigr] d(\hat \eta)  \\
&& = \int_{X} \Bigl[ \frac{P(x|1)}{\hat \eta'}P(1)I_1( \eta)+\frac{P(x|-1)}{\hat \eta'}P(-1)I_{-1}( \eta) \Bigr](\hat \eta'dx)) \nonumber\\
&& = \int_{X} \Bigl[ P(x|1)P(1)I_1( \eta)+P(x|-1)P(-1)I_{-1}( \eta)\Bigr] dx \nonumber\\
&& = \int_{X} P_X(x) \sum_y P(y|x)I_y(\eta) dx \nonumber\\
&& =	\int_{X} P_X(x) J(\eta) dx \nonumber\\
&& =	\int_{X} P_X(x) J(P(1|x)) dx. \nonumber
\end{eqnarray}
The above formulation shows that the distribution of forecasts $s(\hat \eta)$ in the original refinement formulation (\ref{eq:RefOrig})  reduces to $P_X(x)$ which is the distribution of the data. This means that the refinement of a forecaster has nothing to do with how good the forecaster is but depends on the distribution of the underlying data itself which is outside the control of the forecaster. Given observations $x$ the best a forecaster can do is be calibrated. This can also be seen by noting that  refinement   is  a constant term independent of the forecaster  predictions $\hat \eta$ and only  depends on the distribution of the data. This observation leads us to make a connection with the Bayes rule in decision theory which we  explore in the next section.



\subsection{ Refinement and the Bayes rule}
We can think of a forecaster as a kind of classifier that tries to classify days into rainy or sunny. We again assume that the forecaster/classifier has access to a set of observations $x$, for example air pressure. What is the optimal decision a forecaster can make? The Bayes rule tells us that the optimal decision is to choose rainy if $P(1|x)>P(-1|x)$ and sunny otherwise; or equivalently the forecasters predictions should be chance of rain is $\hat \eta=P(1|x)$. This, by definition, is simply the requirement of a calibrated forecaster $\hat \eta=P(1|x)=\eta$.

This can also be written as choose rainy if $\frac{P(x|1)P(1)}{P(x)} > \frac{P(x|-1)P(-1)}{P(x)}$. Assuming no prior knowledge of the chance of rain on any given day we can write choose rainy if $P(x|1) > P(x|-1)$. We see that the optimal decision rule   depends only on the distribution of the data  $P(x|y)$. Given that two forecasters have access to the same air pressure readings, the best forecast they can each give on any given day  depends on the distributions of $P(x|1)$ and $P(x|-1)$ and is simply  $\hat \eta=P(x|1)$. 
Given equal access to data $x$,  both forecasters will make identical predictions. A central part of Bayes decision theory is the Bayes error. We will return to the subject of Bayes error and its connections to refinement in  Section-\ref{sec:FurtherUnderlyData}.


\subsection{Clarifying the refinement concept}
At this point, given the three arguments above, it is evident that the concept of refinement can only be meaningful when comparing forecasters that use different types of evidence or data to form their predictions.
For example, it could be such that one forecaster uses $x_1$ air pressure and another uses an unrelated type of data such as $x_2$ water temperature to make their predictions. In this case both forecasters can be calibrated but one can be more refined than the other since their data distributions $P(x_1)$ and $P(x_2)$ are different.

In summary, for a fixed data type  $x_1$ with distribution $P(x_1)$, the best forecaster possible is the calibrated forecaster and all other forecasters that base their predictions on this type of data $x_1$ can at best be identical to the calibrated forecaster. The only  way to improve on the forecaster's predictions is to use a different type of data or feature $x_2$ with a different distribution of $P(x_2)$ resulting in a calibrated forecaster that has higher refinement. This brings us to the notion of feature selection and its connections to refinement which we explore in the next section.


\section{Refinement, Maximum Marginal Diversity And Conditional Entropy }
\label{sec:RefCondEntropy}
In this section we show that conditional entropy and maximum margin diversity \cite{NUNOMaxDiversityNIPS,NUNOMaxDiversityCVPR} are both special cases of
the extended concept of refinement in the underlying data distribution setting when considering the logistic maximal reward function $J(\eta)=\eta\log(\eta)+(1-\eta)\log(1-\eta)$.

\subsection{Refinement and Maximum Marginal Diversity}
The principal of maximum marginal diversity \cite{NUNOMaxDiversityNIPS,NUNOMaxDiversityCVPR} is studied in feature selection and  states that for a classification problem with observations drawn from a random variable ${\bf Z} \in Z$ and a feature transformations $T_i:Z \rightarrow X_i$, the best feature transformation is the one that leads to a set of maximally diverse marginal densities where the marginal diversity for each feature is defined as
\begin{eqnarray}   
{\bf md}(X_i)=\sum_{y=\{1,-1\}} P_Y(y) D_{KL}(P(x_i|1)||P(x_i)).
\end{eqnarray}
In other words the best feature to use for classification is the one that has the highest ${\bf md}(X_i)$.
Choosing a feature $x_i$ with maximally diverse marginal density is equivalent to choosing a feature with the highest refinement using the logistic $J(\eta)$ function.
This can be shown by writing  (\ref{eq:RefDataSetting}) as 
\begin{eqnarray}
&&S_{Refinement} = \\
&&=\int_{X} \Bigl[ P(x|1)P_Y(1)I_1(P(1|x))+P(x|-1)P_Y(-1)I_{-1}(P(1|x))\Bigr] dx \nonumber\\
&&=\int_{X} \Bigl[ P(x|1)P_Y(1)I_1(\frac{P(x|1)P_Y(1)}{P(x)})+P(x|-1)P_Y(-1)I_{-1}(\frac{P(x|1)P_Y(1)}{P(x)})\Bigr] dx. \nonumber 
\end{eqnarray}


For the special case where $J(\eta)=\eta\log(\eta)+(1-\eta)\log(1-\eta)$ such that $I_1(\eta)=\log(\eta)$ and $I_{-1}(\eta)=\log(1-\eta)$ we have
\begin{eqnarray}
&& S_{Refinement} =  \\ 
&& = \int_{X} \Bigl[ P(x|1)P_Y(1)\log(\frac{P(x|1)P_Y(1)}{P(x)})+P(x|-1)P_Y(-1)\log(1-\frac{P(x|1)P_Y(1)}{P(x)})\Bigr] dx \nonumber\\
&& =P_Y(1)D_{KL}(P(x|1)||P(x)) + P_Y(-1)D_{KL}(P(x|-1)||P(x))  \nonumber \\ 
&&+ P_Y(1)\log(P_Y(1)) + P_Y(-1)\log(P_Y(-1)).  \nonumber
\end{eqnarray}

Assuming that $P_Y(1)=\gamma$ we can write
\begin{eqnarray}
\label{eq:RefKL}
S_{Refinement} \!\!\!\!\!\!\!\!&&=  P_Y(1)D_{KL}\left(P(x|1)||P(x)\right) + P_Y(-1)D_{KL}\left(P(x|-1)||P(x)\right)   \\ 
&&+ P_Y(1)\log(P_Y(1)) + P_Y(-1)\log(P_Y(-1)) \nonumber \\
&&={\bf md}(x)+\gamma \log(\gamma)+ (1-\gamma)\log(1-\gamma) \nonumber
\end{eqnarray}
and maximum marginal diversity  is equivalent, up to a constant, to the refinement formula for the special case of when  $J(\eta)=\eta\log(\eta)+(1-\eta)\log(1-\eta)$. 
The consequences of realizing such an equivalence is that in the case of probability elicitation one realizes that the best a forecaster can do using a certain feature such as $x=$ air pressure is to be calibrated, increased refinement can only come from using better features such as maybe $x=$ air humidity. The insight gained in terms of feature selection is that the KL-divergence is not unique and that other valid $J(\eta)$ functions such as those in Table-\ref{tab:JTableParameteres} and plotted in Figure-\ref{fig:JTableParamPlot} and Figure-\ref{fig:PlotPolinomialJ} can be used to find refinement formulations as seen in Table-\ref{tab:RefFormulasDiffJ} and Table-\ref{tab:RefFormulasDiffJPolyn}. The question that still remains is how  different convex $J(\eta)$ differ in terms of their feature selection properties. We consider this problem in  Sections-\ref{sec:FurtherUnderlyData} and \ref{sec:TighterBounds} .

\subsection{Refinement, mutual information and conditional entropy}
Refinement also has a close relationship with mutual information and conditional entropy. From (\ref{eq:RefKL}) we  write refinement for the special case of $J(\eta)=\eta\log(\eta)+(1-\eta)\log(1-\eta)$ as
\begin{eqnarray}
S_{Refinement} \!\!\!\!\!\!\!\!&&= P_Y(1)D_{KL}(P(x|1)||P(x)) + P_Y(-1)D_{KL}(P(x|-1)||P(x))   \\ 
&&+ P_Y(1)\log(P_Y(1)) + P_Y(-1)\log(P_Y(-1)) \nonumber \\
&&= \sum_{y} P_Y(y)D_{KL}(P(x|y)||P(x)) + \sum_{y} P_Y(y)\log(P_Y(y)) \nonumber \\
&&= I(x;y) +  \sum_{y} P_Y(y)\log(P_Y(y)) \nonumber \\
&&= ( H(y)-H(y|x) ) -H(y) = -H(y|x)  \nonumber
\end{eqnarray}
where $I(x;y)$ is the mutual information and $H(y|x)$ is the conditional entropy. This shows that conditional entropy is a special case of the refinement score when the logistic $J(\eta)$ is used. Note that a higher refinement is a number that is less negative which corresponds to a lower conditional entropy. In other words if  $y$ is completely determined by $x$ then the conditional entropy will be zero, which corresponds to maximum refinement.

Refinement can be directly used for feature selection and is  closely related to conditional mutual information or conditional entropy conditioned on two or more variables. Ranking all  features by their refinement score is not very useful because this does not take into account the dependencies that exist between the features.
Simply using the first $n$ highest ranked features is usually a bad idea since most of the first few features will be redundant, related and dependent. We would like to choose the second feature $z$ such that it not only provides information for classifying the class $y$, but is also complementary to the previously chosen feature $x$. This can be accomplished  by considering  the conditional refinement score defined as
\begin{eqnarray}
S_{Conditional Refinement} =  \sum_{x,z} P(x,z) J(P(1|x,z)).  
\end{eqnarray}
%
Conditional  entropy is a special case of  conditional refinement when the logistic  $J(\eta)=\eta\log(\eta)+(1-\eta)\log(1-\eta)$ is used
\begin{eqnarray}
&& S_{Conditional Refinement}   = \sum_{x,z} P(x,z) J(P(1|x,z))  \\
&&= \sum_{x,z} P(x,z) [P(1|x,z)\log(P(1|x,z)) + (1-P(1|x,z))\log(1-P(1|x,z)) ]  \nonumber \\
&&= \sum_{x,z} P(x,z) \sum_{y} P(y|x,z) \log(P(y|x,z))   \nonumber \\
&&= \sum_{x,z,y} P(x,z,y) \log(P(y|x,z))  \nonumber  \\
&&= -H(y|x,z). \nonumber
\end{eqnarray}

A more practical formula for   conditional refinement  can be written as
\begin{eqnarray}
&& S_{Conditional Refinement}  = \sum_{x,z} P(x,z) J(P(1|x,z))  \\
&&= \sum_{x,y,z} P(z|x,y)P(x|y)P(y)\log(P(y|x,z)) \nonumber \\
&&= \sum_{x,y,z} P(z|x,y)P(x|y)P(y)\log( \frac{P(z|x,y)P(x|y)P(y)}{P(z|x)P(x)}) \nonumber 
\end{eqnarray}
where we have used the logistic $J(\eta)$ to demonstrate. The above formula iteratively picks the best feature
conditioned on the previously chosen feature. Note that all the distributions can be estimated with one dimensional histograms. Conditional entropy has been successfully used in \cite{FastBinaryFeatureSelection} as the basis of a  feature selection algorithm that has been shown to outperform boosting \cite{freund, friedman} and other classifiers on the datasets considered. Finally, note that similar to refinement, different conditional refinement scores can be derived for different choices of convex $J(\eta)$. 

%
%

\section{Refinement And Calibrated Classifiers}
\label{sec:RefCaliClass}
Probability elicitation and  classification by way of conditional risk minimization are closely related and have been most recently  studied in \cite{friedman,Zhang,Buja,HamedNunoLossDesign,Reid}.  A classifier $h$ maps a feature  $x\in  X$ to a class 
label $y \in \{-1,1\}$. This mapping can be written as 
$h(x) = sign[p(x)]$ for a classifier predictor function $p: { X} \rightarrow \mathbb{R}$. 
A predictor function is called an optimal predictor $p^*(x)$ if it minimizes the risk 
\begin{eqnarray}
\label{eq:PredRisk}
R(p)=E_{X,Y}[L(p(x),y)]
\end{eqnarray}
for a given loss $L(p(x),y)$. This is equivalent to minimizing the conditional risk $E_{X|Y}[L(p(x),y)|X=x]$ for all $x$.
Classification can be related to probability elicitation by expressing the predictor as a composite of two functions
\begin{eqnarray}
p(x)=f(\hat \eta(x))
\end{eqnarray}
where $f: [0,1] \rightarrow \mathbb{R}$ is called the {\it link function\/}. The problem of finding the predictor function is now equivalent to finding the link 
and  forecaster functions. A link function is called an optimal link function $f^*(\eta)$ if it is a one-to-one mapping and also implements the Bayes decision rule, meaning that it must be such that
\begin{equation}
  \left\{
  \begin{array}{cc}
    f^* > 0 & \mbox{if $\eta(x) > \frac{1}{2} $} \\
    f^* = 0 & \mbox{if $\eta(x) =  \frac{1}{2} $} \\
    f^* < 0 & \mbox{if $\eta(x) <  \frac{1}{2} $}. 
  \end{array}
  \right.
  \label{eq:Bayesnec}
\end{equation}
Examples of optimal link functions include $f^*=2\eta-1$ and $f^*=\log\frac{\eta}{1-\eta}$, where we have omitted the dependence on $x$ for simplicity. 

A predictor is denoted \emph{calibrated} \cite{DeGroot, Platt, Caruana, Raftery} if it is optimal, i.e. minimizes the risk of (\ref{eq:PredRisk}), and
an optimal link function exists such that 
\begin{eqnarray}
\eta(x)=(f^*)^{-1}(p^*(x))=\hat \eta(x).
\end{eqnarray} 
The  loss $L(p(x),y)$ associated with a calibrated predictor is called a \emph{proper} loss function.

In a classification algorithm a proper loss function is usually fixed beforehand. The associated conditional risk is 
\begin{eqnarray}
C_L(\eta,f)=\eta L(f,1) + (1-\eta)L(f,-1),
\end{eqnarray}
the optimal link function is typically found from
\begin{eqnarray}
f^*_L(\eta)=\arg\min_{f} C_L(\eta,f)
\end{eqnarray}
and the minimum conditional risk is 
\begin{eqnarray}
C^*_L(\eta)=C_L(\eta,f^*_L).
\end{eqnarray}
For example, in the case of the zero-one loss 
\begin{eqnarray}
L_{0/1}(f,y) = \left\{ \begin{array}{ll}
         0, & \mbox{if $y=sign(f)$};\\
        1, & \mbox{if $y \ne sign(f)$},\end{array} \right.
\end{eqnarray}
the associated conditional risk is
\begin{eqnarray}
  C_{0/1}(\eta,f) =  \left\{ \begin{array}{ll}
         1-\eta, & \mbox{if $f \geq 0 $};\\
        \eta, & \mbox{if $f<0$},\end{array} \right.
\end{eqnarray}
the optimal link can be $f^*=2\eta-1$ or $f^*=\log\frac{\eta}{1-\eta}$ and the minimum conditional risk is
\begin{eqnarray}
C^*_{0/1}(\eta)=\min\{\eta,1-\eta\}.
\end{eqnarray}

Margin losses are a special class of loss functions commonly used in classification algorithms which are in the form of 
\begin{eqnarray}
L_{\phi}(f,y)=\phi(yf).
\end{eqnarray}
Margin loss functions assign a non zero penalty to positive $yf$ called the margin.  Algorithms such as boosting \cite{freund, friedman} are based on proper margin loss functions and have not surprisingly demonstrated superior performance given their consistency with the Bayes optimal decision rule \cite{friedman,Buja,HamedNunoLossDesign}.   
Table-\ref{tab:lossesTable} includes some examples of proper margin losses along with their associated optimal links and minimum conditional risks.

\begin{table}[t]
  \centering
  \caption{\protect\footnotesize{Proper margin loss $\phi(v)$, optimal link $f^*_{\phi}(\eta)$,
      optimal inverse link $(f^*_{\phi})^{-1}(v)$  
      and maximal reward  $J(\eta)$.}}
  \resizebox{\textwidth}{!}{ 
  \begin{tabular}{|c|c|c|c|c|}
    \hline
    Loss & $\phi(v)$ & $f^*_{\phi}(\eta)$ & $(f^*_{\phi})^{-1}(v)$ & $J(\eta)$ \\ 
    \hline
    LS\cite{Zhang} & $\frac{1}{2}(1-v)^2$ & $2\eta - 1$ & $\frac{1+v}{2}$ & $-2\eta(1-\eta)$\\
    Exp\cite{Zhang} & $\exp(-v)$ & $\frac{1}{2} \log \frac{\eta}{1-\eta}$& $\frac{e^{2v}}{1+e^{2v}}$ & $-2 \sqrt{\eta (1-\eta)}$\\
    Log\cite{Zhang} & $\log(1+e^{-v})$ & $\log \frac{\eta}{1-\eta}$ & $\frac{e^{v}}{1+e^{v}}$ & $\eta \log \eta + (1-\eta) \log (1 - \eta)$\\
    Savage\cite{HamedNunoLossDesign} & $\frac{4}{(1+e^{v})^2}$ & $\log \frac{\eta}{1-\eta}$ & $\frac{e^{v}}{1+e^{v}}$ & $-4\eta(1-\eta)$\\
    Tangent\cite{HamedNunoTangent} & $(2\arctan(v)-1)^2$ & $\tan(\eta-\frac{1}{2})$ & $\arctan(v)+\frac{1}{2}$ & $-4\eta(1-\eta)$\\
    \hline
  \end{tabular}
  }
  \label{tab:lossesTable}
\end{table}

The score functions $I_1$, $I_{-1}$ and maximal reward function $J(\eta)$ can be related to proper margin losses and the minimum conditional risk by considering the following theorem \cite{HamedNunoLossDesign} which states that if $J(\eta)$ defined as in (\ref{eq:DefJ}) is such that 
\begin{equation}
J(\eta) = J(1-\eta)
\end{equation}
and a continuous function  $f^*_{\phi}(\eta)$ is invertible with symmetry
\begin{equation}
(f^*_{\phi})^{-1}(-v) = 1 -  (f^*_{\phi})^{-1}(v),
\end{equation}
then the functions $I_1$ and $I_{-1}$ derived from
(\ref{eq:I1Jprime}) and (\ref{eq:I2Jprime}) satisfy the following equalities
  \begin{eqnarray}
    I_1(\eta) &=& -\phi(f^*_{\phi}(\eta)) \label{eq:I1f}\\
    I_{-1}(\eta) &=& -\phi(-f^*_{\phi}(\eta)) \label{eq:I-1f},
  \end{eqnarray}
  with
  \begin{equation}
    \phi(v) =  -J\{(f^*_{\phi})^{-1}(v)\} - (1- (f^*_{\phi})^{-1}(v)) J^\prime\{(f^*_{\phi})^{-1}(v)\}.
    \label{eq:phieq}
  \end{equation}
An important direct result of the above theorem is that $J(\eta)=-C^*_{\phi}(\eta)$.
   
The above discussion connects refinement to the classification setting and we can write  refinement  in terms of the calibrated classifier outputs $v=f(\eta)$.
Specifically, assuming a calibrated classifier based on a proper loss function we have 
\begin{equation} 
v=p(x)=f(\eta(x))=f^*(\eta(x))
\end{equation}
and
\begin{equation} 
\hat \eta(x)=\eta(x)=(f^*)^{-1}(v),
\end{equation}
and the refinement term can be written as
\begin{eqnarray}
\label{eq:RefClassSetting}
S_{Refinement}&&=\int_{\hat \eta} s(\hat \eta)J(\eta)d(\hat \eta) \\
&&=\int_{\eta} s(\eta)J(\eta)d(\eta) \nonumber\\
&&=\int_{(f^*_{\phi})^{-1}(v)} s((f^*_{\phi})^{-1}(v))J((f^*_{\phi})^{-1}(v))d((f^*_{\phi})^{-1}(v)) \nonumber\\
&&=\int_v \frac{s(v)}{((f^*_{\phi})^{-1}(v))'}J((f^*_{\phi})^{-1}(v)) ((f^*_{\phi})^{-1}(v))'d(v) \nonumber\\
&&=\int_v s(v)J((f^*_{\phi})^{-1}(v))d(v). \nonumber
\end{eqnarray}

\begin{figure*}[t]
  \centering
     \includegraphics[width=4in]{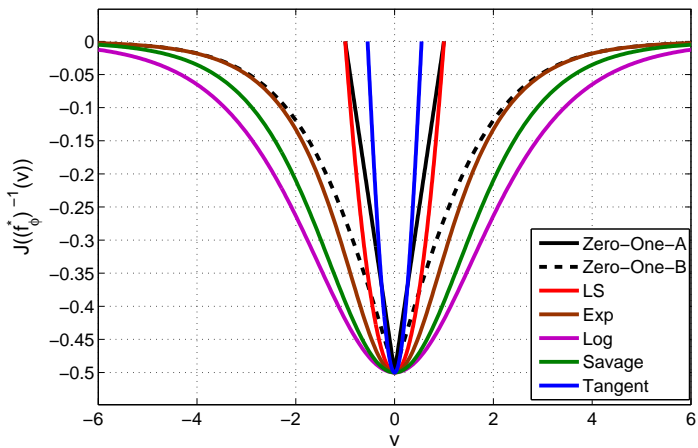} 
 \caption{\protect{\footnotesize{Plot of 
$J((f^*_{\phi})^{-1}(v))$ for different loss functions.}}}
  \label{fig:LogJPlot}
\end{figure*}

For the special case of the proper margin Log loss of Table-\ref{tab:lossesTable} we have
\begin{equation}
J(\eta)=0.7213[\eta\log(\eta)+(1-\eta)\log(1-\eta)]
\end{equation}
and 
\begin{equation}
(f^*_{\phi})^{-1}(v)=\frac{e^v}{1+e^v}=\eta
\end{equation}
so $J((f^*_{\phi})^{-1}(v))=J(\eta)$ can be simplified to
\begin{equation}
J((f^*_{\phi})^{-1}(v))=0.7213\left[\frac{ve^v}{1+e^v}-\log(1+e^v)\right]
\end{equation}
and is plotted in Figure-\ref{fig:LogJPlot}.
The  $J((f^*_{\phi})^{-1}(v))$ functions associated with the proper margin losses of  Table-\ref{tab:lossesTable} are presented in Table-\ref{tab:LossRefFuncs} and plotted in Figure-\ref{fig:LogJPlot}. Refinement  for the log loss can be written as
\begin{eqnarray}
S_{Refinement}=\int_v s(v)J((f^*_{\phi})^{-1}(v))d(v) =0.7213\int_v s(v) \Bigl[ \frac{ve^v}{1+e^v}-\log(1+e^v) \Bigr] d(v) 
\end{eqnarray}
where  we reiterate that $s(v)$ is the distribution of the classifier's predictions.

\begin{table}[t]
  \centering
  \caption{\protect\footnotesize{The $J((f^*_{\phi})^{-1}(v))$, the domain of $v$ over which it is defined and the corresponding $J(\eta)$ and $(f^*_{\phi})^{-1}(v)$.}}
  \resizebox{\textwidth}{!}{ 
  \begin{tabular}{|c|c|c|c|c|}
    \hline
    Loss & $J((f^*_{\phi})^{-1}(v))$ & $(f^*_{\phi})^{-1}(v)$ & $J(\eta)$ & Domain  \\ 
    \hline
    Zero-One-A & $-\min\{\frac{v+1}{2},1-\frac{v+1}{2}\}$ & $\frac{v+1}{2}$ & $-\min\{\eta,1-\eta\}$ & $[-1~1]$ \\
    Zero-One-B & $-\min\{\frac{e^v}{1+e^v},1-\frac{e^v}{1+e^v}\}$ & $\frac{e^v}{1+e^v}$ & $-\min\{\eta,1-\eta\}$ & $[-\infty~\infty]$ \\
    LS & $\frac{1}{2}(v^2-1)$ & $\frac{v+1}{2}$ & $-2\eta(1-\eta)$ & $[-1~1]$ \\
    Exp & $-\sqrt{\frac{e^{2v}}{(1+e^{2v})^2}}$ & $\frac{e^{2v}}{1+e^{2v}}$ & $-\sqrt{\eta(1-\eta)}$ & $[-\infty~\infty]$ \\
    Log & $0.7213[\frac{ve^v}{1+e^v}-\log(1+e^v)]$ & $\frac{e^v}{1+e^v}$ & $0.7213[\eta\log(\eta)+(1-\eta)\log(1-\eta)]$ & $[-\infty~\infty]$\\
    Savage & $\frac{-2e^v}{(1+e^v)^2}$ & $\frac{e^v}{1+e^v}$ & $-2\eta(1-\eta)$ & $[-\infty~\infty]$\\
    Tangent & $2(\arctan(v))^2-\frac{1}{2}$ & $\arctan(v)+\frac{1}{2}$ & $-2\eta(1-\eta)$ & $[-\tan(\frac{1}{2})~\tan(\frac{1}{2})]$\\
    \hline
  \end{tabular}
  }
  \label{tab:LossRefFuncs}
\end{table}

All of the plotted $J((f^*_{\phi})^{-1}(v))$ functions in Figure-\ref{fig:LogJPlot} are quasi convex. This is shown to be always the case by considering
the derivative 
\begin{eqnarray}
\frac{\partial J((f^*_{\phi})^{-1}(v))}{\partial v} &&= \frac{\partial J((f^*_{\phi})^{-1}(v))}{\partial (f^*_{\phi})^{-1}(v)} 
\frac{\partial (f^*_{\phi})^{-1}(v)}{\partial v}  \\
&&=J'((f^*_{\phi})^{-1}(v)) \frac{\partial (f^*_{\phi})^{-1}(v)}{\partial v} \nonumber
\end{eqnarray}
and the fact that $(f^*_{\phi})^{-1}(v)$ is a nondecreasing invertible function and $J(\eta)$ is convex. 
Since $\frac{\partial (f^*_{\phi})^{-1}(v)}{\partial v}>0$ and $J'((f^*_{\phi})^{-1}(v))$ changes sign only once, the derivative of $J((f^*_{\phi})^{-1}(v))$ also changes sign only once proving that $J((f^*_{\phi})^{-1}(v))$ is quasi convex.

Given the quasi convex shape of $J((f^*_{\phi})^{-1}(v))$, the refinement of a classifier increases when  the distribution of the classifier predictions $s(v)$
is concentrated away from the decision boundary. A classifier with predictions that are concentrated further away from the boundary is preferable and the refinement of a classifier can be thought of as a measure of the \emph{classifier's marginal density}. This observation is formally considered in Section-\ref{sec:FurtherClassOut} and  allows for the comparison of calibrated  classifiers based solely on the distribution of their predictions. 

We note that although the concept of classifier marginal density seems to be related to maximum margin theory \cite{book:vapnik} in classifier design, there are a few key difference. 1) Calibrated classifiers that are built from the same underlying data distribution will have the same classifier marginal density but can have different margins. The notion of margins is thus in contradiction to the axioms of probability theory, while the concept of classifier marginal density is not. 2) margins are only defined for completely separable data, while classifier marginal density does not have such restrictions. 3) While the margin of a classifier considers  only the data that lie on the margin, the notion of classifier marginal density considers the entire spread and distribution of the data.

\section{Further Insight Into refinement In The Original Probability Elicitation Setting}
\label{sec:FurtherOrigElic}
The original refinement formulation is in the probability elicitation setting and was formulated as 
\begin{eqnarray}
\label{eq:RefOrig2}
S_{Refinement}=\int_{\hat \eta} s(\hat \eta)J(\eta) d(\hat \eta)
\end{eqnarray}
in Section-\ref{sec:RefElic}. As mentioned previously, it is intuitive that refinement increases as the distribution of the predictions $s(\hat \eta)$  concentrates around $\hat\eta=0$ and $\hat \eta=1$. We formalize this intuition in this section and derive the maximum and minimum refinement scores using an inner product Hilbert space interpretation. 

Real continuous functions $f(x)$ and $g(x)$ that are also square integrable  form an inner product Hilbert space \cite{book:OptVector} where the inner product is defined as
\begin{eqnarray}
<f,g>=\int f(x)g(x) d(x)
\end{eqnarray} 
with induced norm of
\begin{eqnarray}
||f||^2=\int \left( f(x) \right)^2 d(x).
\end{eqnarray} 
Let $\int \left|J(\eta) \right|^2 d(\eta) < \infty$ and $\int \left|s(\hat \eta) \right|^2 d(\hat \eta) < \infty$ then $J(\eta)$ and $s(\hat \eta)$ are square integrable functions. The inner product associated with this inner product Hilbert space is
\begin{eqnarray}
<s,J>=\int_{\hat \eta} s(\hat \eta)J(\eta) d(\hat \eta).
\end{eqnarray} 
which is equal to the original refinement formulation of (\ref{eq:RefOrig2}). In other words refinement computes the inner product between the two functions
$J(\eta)$ and $s(\hat \eta)$.  As seen in Table-\ref{tab:lossesTable} $J(\eta) \le 0$ and $s(\hat \eta) \ge 0$ since $s(\hat \eta)$ is a probability distribution function. This constrains the refinement score to $<s,J> \le 0$. The maximum and minimum refinement scores for a fixed $J(\eta)$ can now be computed by considering the inner product between a fixed $J(\eta)$ and a distribution of prediction functions $s(\hat \eta)$.
Specifically the minimum refinement score is 
\begin{eqnarray}
S^{Min}_{Refinement}=<s,J>=||s|| \cdot ||J|| \cdot \cos(\theta)= \alpha ||J|| \cdot ||J|| \cdot (-1) = -\alpha||J||^2
\end{eqnarray} 
and corresponds to when $s=-\alpha J$ for some multiple $\alpha$.   
The maximum refinement  score is 
\begin{eqnarray}
S^{Max}_{Refinement}=<s,J>=||s|| \cdot ||J|| \cdot \cos(\theta)=||s|| \cdot ||J|| \cdot (0) = 0
\end{eqnarray} 
and corresponds to when $s \perp J$.

\begin{figure*}[t]
  \centering
     \includegraphics[width=2.4in]{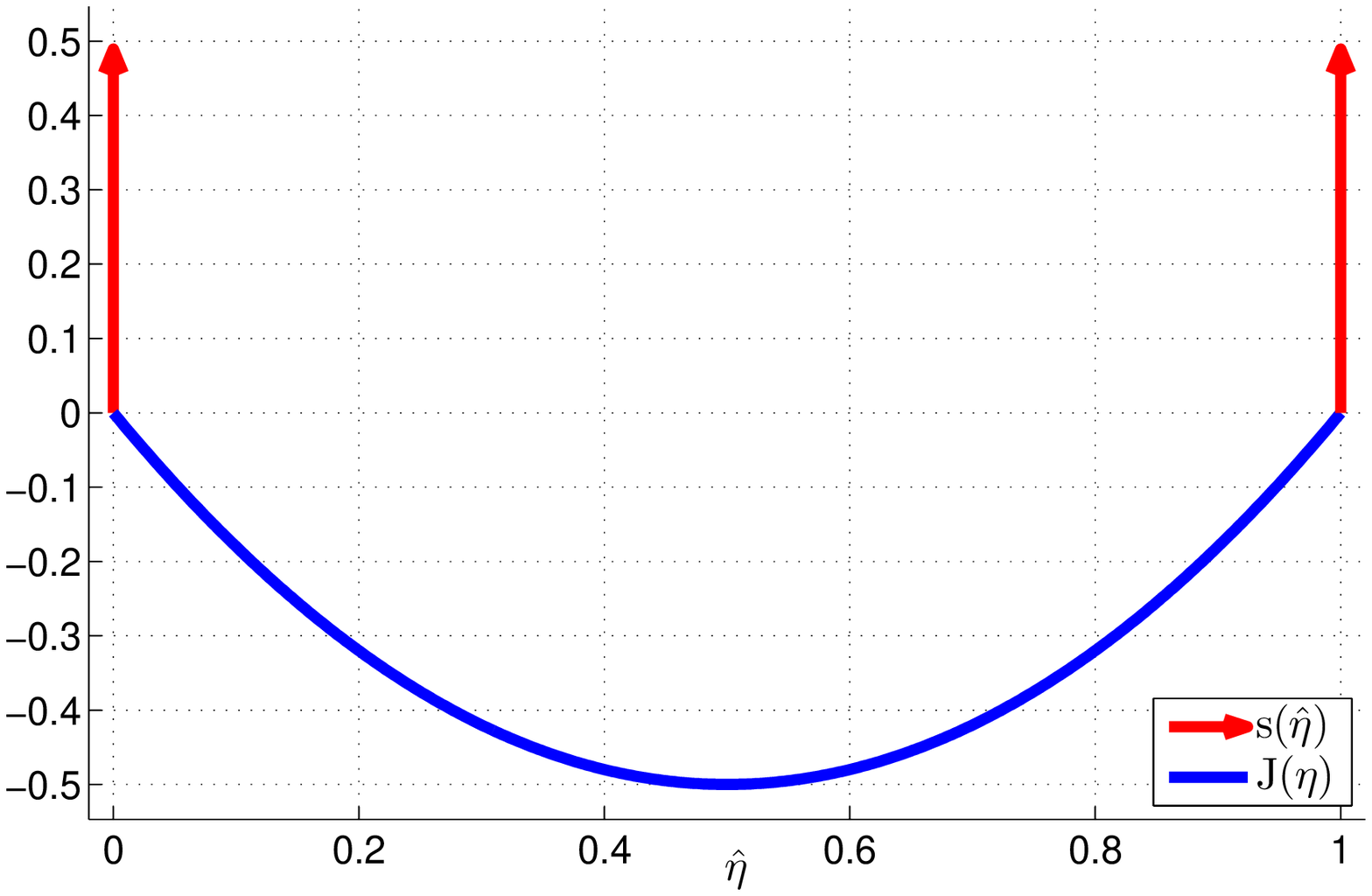} 
    \includegraphics[width=2.4in]{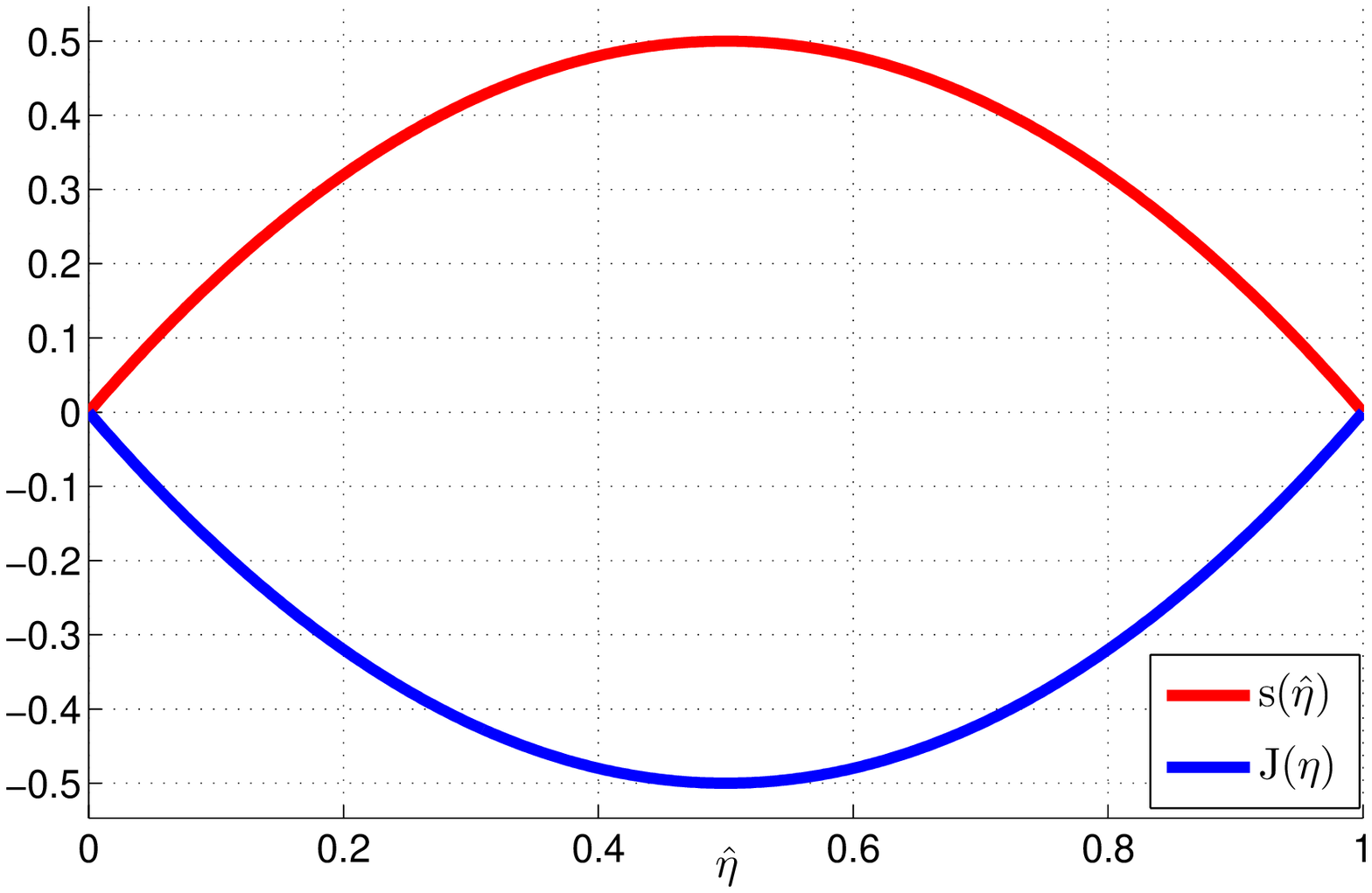} 
 \caption{\protect{\footnotesize{ Plot of $s(\hat \eta)$ for maximum (left) and minimum (right) refinement  with a hypothetical $J(\eta)$.}}}
  \label{fig:MaxMinRefPlot}
\end{figure*}

Usually, the score functions  $I_1$ and $I_{-1}$ are chosen to be symmetric such that $I_1(\eta)=I_{-1}(1-\eta)$ so that the scores attained for 
predicting either class  $y=\{1,-1\}$ remain class insensitive. In this case the corresponding $J(\eta)$ is also symmetric such that $J(\eta)=J(1-\eta)$.
This can be confirmed by noting that
\begin{eqnarray}
J(1-\eta)&=&(1-\eta)I_1(1-\eta)+(1-1+\eta)I_{-1}(1-\eta) \\
&=&(1-\eta)I_{-1}(\eta)+\eta I_1(\eta)=J(\eta). \nonumber
\end{eqnarray}
When $J(\eta) \le 0$ is convex  symmetric over $\eta \in \{0~1\}$ then $J(\eta)$ is minimum at $\eta=\frac{1}{2}$ and $J(0)=J(1)=0$ and the maximum refinement score verifiably corresponds   to when all of the predictions are either $0$ or $1$ such that
$s(\hat \eta)=\gamma\delta(\hat \eta) + (1-\gamma)\delta(1-\hat \eta)$ where $0 \le \gamma \le 1$. 
The $s(\hat \eta)$ pertaining to the cases of maximum and minimum refinement are plotted for a hypothetical symmetric $J(\eta)$ in Figure-\ref{fig:MaxMinRefPlot}.

\section{Further Insight Into Refinement In The Classifier Output Setting}
\label{sec:FurtherClassOut}
In Section-\ref{sec:RefCaliClass} we stated that when considering refinement in the classifier outputs setting under the formulation of (\ref{eq:RefClassSetting}), refinement increases as $s(v)$ is concentrated away from the boundary. This can be formally addressed by letting $s(v)$ and $J((f^*_{\phi})^{-1}(v))$ be square integrable functions that form an inner product Hilbert space with inner product 
\begin{eqnarray}
\label{eq:RefClassSetting2}
<s,J>=\int_v s(v)J((f^*_{\phi})^{-1}(v))d(v).
\end{eqnarray}
This is equal to the refinement formulation of (\ref{eq:RefClassSetting}) associated with the classifier output setting.
An argument similar to that of Section-\ref{sec:FurtherOrigElic} leads  to the conclusion that refinement in the classifier output setting
is minimum when the distribution of classifier outputs $s(v)=-\alpha J((f^*_{\phi})^{-1}(v))$, increases as $s(v)$ concentrates away from the decision boundary
and is maximum when $s(v)=\lim_{t \to \infty}\gamma\delta(v-t) + (1-\gamma)\delta(v+t)$ where $0 \le \gamma \le 1$.


\section{Further Insight Into Refinement In The Underlying Data Distribution Setting}
\label{sec:FurtherUnderlyData}
In Section-\ref{sec:ExtendRefToUnderlyData} we showed that the refinement score can be reduced to the underlying data distribution setting as
\begin{eqnarray}
S_{Refinement}= 	\int_{X} P_X(x) J(P(1|x)) dx.
\end{eqnarray}
Here we expand on this formulation and formalize its connections to the Bayes error and eventually derive  novel measures that provide arbitrarily tighter
bounds on the Bayes error.

First we show that refinement in the data distribution setting is also an inner product Hilbert space with inner product defined as 
\begin{eqnarray}
\label{eq:RefInnerUnderlyDataDist}
<P_X,J> = 	\int_{X} P_X(x) J(P(1|x)) dx.
\end{eqnarray}
This follows directly from letting $P_X(x)$ and $J(P(1|x))$ be square integrable functions which is not a stringent constraint since most probability density functions  are square integrable \cite{book:WaveletTheory}.
We  also directly show that $J(P(1|x)) \le 0$ is quasi convex over $x$. This follows from 
\begin{eqnarray}
\frac{\partial J(P(1|x))}{\partial x} &&= \frac{\partial J(P(1|x))}{\partial P(1|x)} 
\frac{\partial P(1|x)}{\partial x}   \\
&&=J'(P(1|x)) \frac{\partial P(1|x)}{\partial x}, \nonumber
\end{eqnarray}
the fact that $\eta(x)=P(1|x)$ is an  invertible and hence monotonic function from (\ref{eq:invrtPeta}) and $J(\eta)$ is convex. 
Since $\frac{\partial P(1|x)}{\partial x}>0~\forall x$ or $\frac{\partial P(1|x)}{\partial x}<0~\forall x$ and $J'(P(1|x))$ changes sign only once, the derivative of $J(P(1|x))$ also changes sign only once proving that $J(P(1|x))$ is quasi convex.

Once again, refinement  in the data distribution setting $<P_X,J>\le 0$ is minimum when $P_X(x)=-\alpha J(P(1|x))$, and is maximum and equal to zero when $P_X(x) \perp J(P(1|x))$. 

Assuming equal priors $P(1)=P(-1)=\frac{1}{2}$,
\begin{eqnarray}
P(x)=\frac{P(x|1)+P(x|-1)}{2}
\end{eqnarray}
and 
\begin{eqnarray}
P(1|x)=\frac{P(x|1)}{P(x|1)+P(x|-1)}.
\end{eqnarray}
We can write refinement in terms of the underlying data distributions $P(x|1)$ and $P(x|-1)$ as
\begin{eqnarray}
S_{Refinement}=  
	\int_{X} (\frac{P(x|1)+P(x|-1)}{2}) J(\frac{P(x|1)}{P(x|1)+P(x|-1)}) dx. 
\end{eqnarray}

\begin{figure*}[t]
  \centering
     \includegraphics[width=5in]{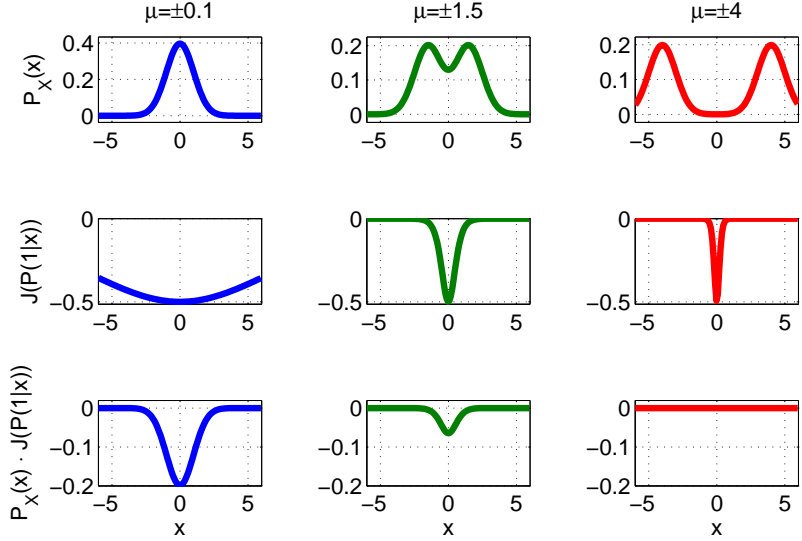} 
 \caption{\protect{\footnotesize{Plot of the $J_{LS}$ refinement terms for three different unit variance Gaussians.}}}
  \label{fig:RefinementTermsThreeGauss}
\end{figure*}

For example under the least squares $J_{LS}(P(1|x))=2P(1|x)(P(1|x)-1)$, the refinement formulation simplifies to
\begin{eqnarray}
S^{LS}_{Refinement}=\int \frac{-P(x|1)P(x|-1)}{(P(x|1)+P(x|-1))} dx.
\end{eqnarray}
Plot-\ref{fig:RefinementTermsThreeGauss} shows the   $P(x)$, $J(P(1|x))$ and $P(x)J(P(1|x))$ terms for three Gaussian distributions of unit variance and means of $\mu=\pm 0.1$, $\mu=\pm 1.5$ and $\mu=\pm 4$. In accordance with the inner product interpretation, as the means separate and the two distributions $P(x|1)$ and $P(x|-1)$ have less  overlap, the refinement increases (is less negative) and approaches zero. 

\begin{figure*}[t]
  \centering
     \includegraphics[width=4in]{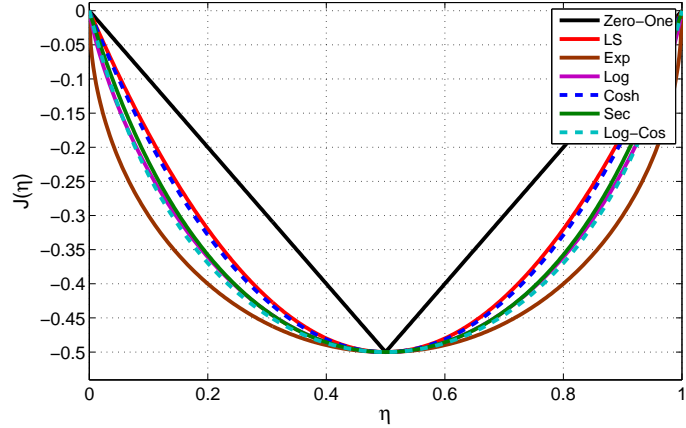} 
 \caption{\protect{\footnotesize{Plot of the $J(\eta)$ in Table-\ref{tab:JTableParameteres}.}}}
  \label{fig:JTableParamPlot}
\end{figure*}

In Table-\ref{tab:RefFormulasDiffJ} we have derived the refinement formulation  for the different $J(P(1|x))$ of Table-\ref{tab:JTableParameteres} which are plotted in Figure-\ref{fig:JTableParamPlot}. Refinement for the zero-one maximum conditional score function  
\begin{eqnarray}
\label{eq:JzeroOne}
J_{0/1}(\eta)=\left\{ \begin{array}{ll}
         -(1-\eta), & \mbox{if $\eta \geq \frac{1}{2} $};\\
        -\eta, & \mbox{if $\eta<\frac{1}{2}$},\end{array} \right.
\end{eqnarray}
is
\begin{eqnarray}
\label{eq:ZeroOneRefinement}
&&S^{0/1}_{Refinement}=\int P(x)J_{0/1}(P(1|x)) dx= \\
&&\int_{P(1|x)\geq\frac{1}{2}} (\frac{P(x|1)+P(x|-1)}{2})(-(1-\frac{P(x|1)}{P(x|1)+P(x|-1)})) dx + \nonumber\\
&&\int_{P(1|x) < \frac{1}{2}} (\frac{P(x|1)+P(x|-1)}{2})(-(\frac{P(x|1)}{P(x|1)+P(x|-1)})) dx = \nonumber\\
&&-\frac{1}{2}\int_{P(1|x)\geq\frac{1}{2}} P(x|-1) dx -\frac{1}{2}\int_{P(1|x)< \frac{1}{2}} P(x|1) dx =-\frac{1}{2}(\epsilon_1+\epsilon_2)=-\epsilon \nonumber
\end{eqnarray}
where $\epsilon_2$ is the miss rate, $\epsilon_1$ is the false positive rate and $\epsilon$ is the Bayes error rate. In other words, refinement under the zero-one $J_{0/1}(\eta)$ is equal to  minus the Bayes error.
When refinement is computed under the other $J(\eta)$ of Table-\ref{tab:JTableParameteres},  an upper bound on  the Bayes error is  being computed. 
This can be formally written as  
\begin{eqnarray}
\label{eq:JZeroOneDiff}
&&S_{Refinement}^{0/1}-S_{Refinement}^{J(\eta)}= \epsilon - S_{Refinement}^{J(\eta)}\\
&&\int_{x}P_X(x)J_{0/1}(P(1|x))dx - \int_{x}P_X(x)J(P(1|x))dx = \nonumber \\
&& \int_{x}P_X(x) \biggl( J_{0/1}(P(1|x))-J(P(1|x)) \biggr) dx. \nonumber
\end{eqnarray} 
In other words, the $J(\eta)$ that are closer to the $J_{0/1}(\eta)$ result in refinement formulations that provide tighter bounds on the Bayes error.  Figure-\ref{fig:JTableParamPlot} shows that  $J_{LS}$, $J_{Cosh}$, $J_{Sec}$, $J_{Log}$, $J_{Log-Cos}$ and $J_{Exp}$ are in order the closest to  $J_{0/1}$ and the corresponding refinement formulations in Table-\ref{tab:JTableParameteres} provide in the same order tighter bounds on the Bayes error. This can also be directly verified by noting that $S_{Exp}$ is equal to the Battacharyy bound \cite{book:Fukunaga}, $S_{LS}$ is equal to the asymptotic nearest neighbor bound \cite{book:Fukunaga,NNClassification} and $S_{Log}$ is equal to the Jensen–Shannon divergence \cite{JenShannonLin}. These three formulations have been independently studied throughout the literature and the fact that they produce upper bounds on the Bayes error have been directly verified. Here we have rederived these three measures by resorting to the concept of refinement which not only allows us to provide a  unified approach to these different methods but has also led to a systematic method for deriving novel  refinement measures or bounds on the Bayes error, namely the $S_{Cosh}$, $S_{Log-Cos}$ and the $S_{Sec}$.

\begin{table}[tbp]
\centering
\caption{\protect\footnotesize{Refinement measure for different $J(\eta)$ }}
\resizebox{\textwidth}{!}{ 
\begin{tabular}{|c|c|}
    \hline
    $J(\eta)$ & $S_{Refinement}$ \\
    \hline
    Zero-One & Bayes Error \\
    &  \\
    \hline
    LS & $\int_x \frac{-P(x|1)P(x|-1)}{P(x|1)+P(x|-1)} dx$ \\
    &  \\
    \hline
    Exp & $-\frac{1}{2}\int_x \sqrt{P(x|1)P(x|-1)} dx$ \\
    &  \\
    \hline
    Log & $\frac{0.7213}{2}D_{KL}(P(x|1)||P(x|1)+P(x|-1))+\frac{0.7213}{2}D_{KL}(P(x|-1)||P(x|1)+P(x|-1))$ \\
    &  \\
    \hline
    Log-Cos & $\int_x \frac{P(x|1)+P(x|-1)}{2} \left[ \frac{-1}{2.5854}\log\left(\frac{\cos(\frac{2.5854(P(x|1)-P(x|-1))}{2(P(x|1)+P(x|-1))})}{cos(\frac{2.5854}{2})}\right) \right] dx$ \\
    &  \\
    \hline
    Cosh & $\int_x \frac{P(x|1)+P(x|-1)}{2} \left[ \cosh(\frac{1.9248(P(x|-1)-P(x|1))}{2(P(x|1)+P(x|-1))}) -\cosh(\frac{-1.9248}{2}) \right] dx$ \\
    &  \\
    \hline
    Sec & $\int_x \frac{P(x|1)+P(x|-1)}{2} \left[ \sec(\frac{1.6821(P(x|-1)-P(x|1))}{2(P(x|1)+P(x|-1))}) -\sec(\frac{-1.6821}{2}) \right]  dx$ \\
    &  \\
    \hline
\end{tabular}
}
\label{tab:RefFormulasDiffJ}
\end{table}

\begin{table}[tbp]
  \centering
  \caption{\protect\footnotesize{ $J$ specifics used to compute the refinement score.}}
  \begin{tabular}{|c|c|}
    \hline
    Method & $J(\eta)$  \\
    \hline
    LS & $2\eta(\eta-1)$ \\
    \hline
    Log & $0.7213(\eta\log(\eta)+(1-\eta)\log(1-\eta))$ \\
    \hline
    Exp & $-\sqrt{\eta(\eta-1)}$ \\
    \hline
    Log-Cos & $(\frac{-1}{2.5854})\log(\frac{\cos(2.5854(\eta-\frac{1}{2}))}{\cos(\frac{2.5854}{2})})$ \\
    \hline
    Cosh & $\cosh(1.9248(\frac{1}{2}-\eta))-\cosh(\frac{-1.9248}{2})$ \\
    \hline
    Sec &  $\sec(1.6821(\frac{1}{2}-\eta))-\sec(\frac{-1.6821}{2})$ \\
    \hline
  \end{tabular}
  \label{tab:JTableParameteres}
\end{table}

\section{Measures With Tighter Bounds On The Bayes Error}
\label{sec:TighterBounds}
Although the three novel refinement score functions discussed above provide relatively tighter upper bounds on the Bayes error, they do not produce the tightest bounds. In Table-\ref{tab:RefFormulasDiffJ} $J_{LS}(\eta)$ provides the closest approximation to  $J_{0/1}(\eta)$, thus resulting in a  tighter bound. A natural question is if the refinement approach can be used to derive formulations that provide even tighter bounds on the Bayes error. In order to do so, (\ref{eq:JZeroOneDiff}) states that we simply need to  find $J(\eta)$ that are  closer approximations to  $J_{0/1}(\eta)$. In this section we derive   polynomial functions $J_{Poly}(\eta)$ that are arbitrarily close approximations to   $J_{0/1}(\eta)$ thus leading to measures that have the tightest bounds on the Bayes error.

The Weierstrass approximation theorem \cite{book:RealAnalysis, book:NumericalAnalysis} states that for a continuous function $f(x)$ defined on $[a,b]$ 
there exists a polynomial $P(x)$ that is as close to $f(x)$ as desired such that 
\begin{eqnarray}
|f(x)-P(x)|<\epsilon, ~\forall x \in [a,b].
\end{eqnarray}
With $J_{0/1}(\eta)$ as the target function, we  demonstrate a general procedure for deriving a class of polynomial functions $J_{Poly-n}(\eta)$ that are as close to $J_{0/1}(\eta)$ as desired. As an example, we  derive the $J_{Ploy-2}(\eta)$ which leads to the $S_{Poly-2}$ bound on the Bayes error which is a tighter bound on the Bayes error than $J_{LS}(\eta)$. We  also derive the $S_{Poly-4}$ bound which is an even tighter bound and  show that  $J_{LS}=J_{Poly-0}$. 

When $J(\eta)$ is convex symmetric over $\eta \in \{0~1\}$ then $J(\eta)$ is minimum at $\eta=\frac{1}{2}$ and so $J'(\frac{1}{2})=0$. The symmetry $J(\eta)=J(1-\eta)$
results in a similar constrain on the second derivative $J''(n)=J''(1-\eta)$ and convexity requires that the second derivative satisfy $J''(\eta)>0$. 
The symmetry  and convexity constraint can both be satisfied by considering
\begin{eqnarray}
J''_{Poly-n}(\eta)=(\eta(1-\eta))^n
\end{eqnarray}
where $n$ is an even number. From this we  write
\begin{eqnarray}
J'_{Poly-n}(\eta)=\int(\eta(1-\eta))^n d(\eta) + K_1 = Q(\eta)+K_1.
\end{eqnarray}
Satisfying the constraint that $J'_{Poly-n}(\frac{1}{2})=0$, we  find $K_1$ as
\begin{eqnarray}
K_1=- \left. \int(\eta(1-\eta))^n d(\eta) \right |_{\eta=\frac{1}{2}} = -Q(\frac{1}{2}).
\end{eqnarray}
Finally, $J_{Poly-n}(\eta)$ is
\begin{eqnarray}
J_{Poly-n}(\eta)=K_2(\int  Q(\eta) d(\eta) +K_1\eta)=K_2(R(\eta)+K_1\eta),
\end{eqnarray}
where $K_2$ is a scaling factor  such that
\begin{eqnarray}
K_2=\frac{-0.5}{\left. (\int  Q(\eta) d(\eta) +K_1\eta)\right|_{\eta=\frac{1}{2}}}.
\end{eqnarray}
In other words this scaling factor is set to satisfy $J_{Poly-n}(\frac{1}{2})=J_{0/1}(\frac{1}{2})=-\frac{1}{2}$.

As an example, we  derive $J_{Poly-2}$. Following the procedure above 
\begin{eqnarray}
J''_{Poly-2}(\eta)=(\eta(1-\eta))^2=\eta^2+\eta^4-2\eta^3 >0.
\end{eqnarray} 
From this we have
\begin{eqnarray}
J'_{Poly-2}(\eta)=\frac{1}{3}\eta^3+\frac{1}{5}\eta^5-\frac{2}{4}\eta^4 + K_1.
\end{eqnarray}
Satisfying $J'_{Poly-2}(\frac{1}{2})=0$ we find $K_1=-0.0167$. Therefore,
\begin{eqnarray}
J_{Poly-2}(\eta)=K_2(\frac{1}{12}\eta^4 +\frac{1}{30}\eta^6 -\frac{1}{10}\eta^5 +(-0.0167)\eta).
\end{eqnarray}
Satisfying $J_{Poly-2}(\frac{1}{2})=-\frac{1}{2}$ we find $K_2=87.0196$.

\begin{figure*}[t]
  \centering
     \includegraphics[width=4in]{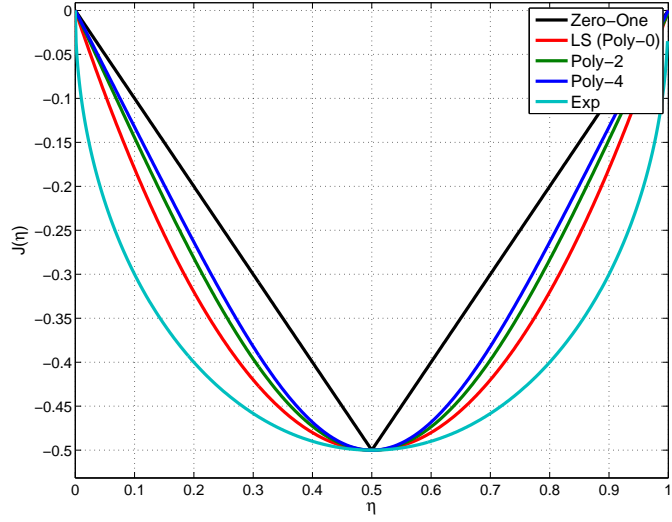} 
 \caption{\protect{\footnotesize{Plot of $J_{Poly-n}(\eta)$.}}}
  \label{fig:PlotPolinomialJ}
\end{figure*}

Figure-\ref{fig:PlotPolinomialJ} plots $J_{Poly-2}(\eta)$ which shows that, as expected, it is a closer approximation to $J_{0/1}(\eta)$ when compared to $J_{LS}(\eta)$. Following the same steps, it is readily shown  that $J_{LS}(\eta)=J_{Poly-0}(\eta)$, meaning that $J_{LS}(\eta)$ is derived from the special case of $n=0$. As we increase $n$, we increase the order of the resulting polynomial which provides a tighter fit to $J_{0/1}(\eta)$. Figure-\ref{fig:PlotPolinomialJ} also plots $J_{Poly-4}(\eta)$ 
\begin{eqnarray}
&&J_{Poly-4}(\eta)= \\
&&1671.3(\frac{1}{90}\eta^{10} -\frac{1}{18}\eta^9 +\frac{3}{28}\eta^8 -\frac{2}{21}\eta^7 +\frac{1}{30}\eta^6 +(-7.9365\times10^{-4})\eta)  \nonumber
\end{eqnarray}
and we see that this provides an even closer approximation to $J_{0/1}(\eta)$. 
Table-\ref{tab:RefFormulasDiffJPolyn} shows the corresponding refinement measure for each of the $J_{Poly-n}(\eta)$ functions, with $S_{Poly-4}$ providing
the tightest bound on the Bayes error.  Arbitrarily tighter bounds are possible by simply using $J_{Poly-n}$ with larger $n$.

\begin{table}[tbp]
\centering
\caption{\protect\footnotesize{Refinement measure for different $J_{Poly-n}(\eta)$ }}
\resizebox{\textwidth}{!}{ 
\begin{tabular}{|c|c|}
    \hline
    $J(\eta)$ & $S_{Refinement}$ \\
    \hline
    Zero-One & Bayes Error \\
    &  \\
    \hline
    Poly-0 (LS) & $\int \frac{-P(x|1)P(x|-1)}{P(x|1)+P(x|-1)} dx$ \\
    &  \\
    \hline
    Poly-2 & $\frac{K_2}{2} \int \frac{P(x|1)^4}{12(2P(x))^3} + \frac{P(x|1)^6}{30(2P(x))^5} -\frac{P(x|1)^5}{10(2P(x))^4} -K_1P(x|1) dx   $ \\
    & $K_1=0.0167,K_2=87.0196,P(x)=\frac{P(x|1)+P(x|-1)}{2}$ \\
    & \\
    \hline
    Poly-4 & $\frac{K_2}{2} \int \frac{P(x|1)^{10}}{90(2P(x))^9} -\frac{P(x|1)^9}{18(2P(x))^8} +\frac{3P(x|1)^8}{28(2P(x))^7}     -\frac{2P(x|1)^7}{21(2P(x))^6} +\frac{P(x|1)^6}{30(2P(x))^5}-K_1P(x|1)  dx$ \\
    & $K_1=7.9365\times10^{-4},K_2=1671.3,P(x)=\frac{P(x|1)+P(x|-1)}{2}$ \\
    & \\
    \hline
\end{tabular}
}
\label{tab:RefFormulasDiffJPolyn}
\end{table}

\section{Conclusion}
\label{sec:conclusion}
The concept of refinement was first established in the probability elicitation literature and despite its close connections to proper scoring functions, has largely remained restricted to the forecasting literature. In this work we have revisited this important statistical measure from the viewpoint of machine learning. In particular, this concept is first considered from a fundamental perspective with the basic axioms of probability. This deeper understanding of refinement is used as a guide to extend refinement from the original probability elicitation setting to two novel formulations namely the underlying data distribution and classifier output settings. These three refinement measures were then shown to be inner products in their respective Hilbert spaces. This unifying abstraction was then used to connect ideas such as maximum marginal diversity, conditional entropy, calibrated classifiers and Bayes error. Specifically we showed that maximal marginal diversity and conditional entropy are special cases of refinement in the underlying data distribution setting and introduced  conditional refinement. Also a number of novel refinement measures were presented for the comparison of classifiers under the classifier output setting. Finally, refinement in the underlying data distribution setting was used in a general procedure for deriving arbitrarily tight bounds on the Bayes error.

\vskip 0.2in
\bibliography{RefinementRevisitedJMLRFormat}

\begin{thebibliography}{36}
\providecommand{\natexlab}[1]{#1}
\providecommand{\url}[1]{\texttt{#1}}
\expandafter\ifx\csname urlstyle\endcsname\relax
  \providecommand{\doi}[1]{doi: #1}\else
  \providecommand{\doi}{doi: \begingroup \urlstyle{rm}\Url}\fi

\bibitem[Bartle(1976)]{book:RealAnalysis}
Robert~G. Bartle.
\newblock \emph{The Elements of Real Analysis (Second Edition)}.
\newblock John Wiley and Sons, New York, 1976.

\bibitem[Bröcker(2009)]{Brocker2009}
Jochen Bröcker.
\newblock Reliability, sufficiency, and the decomposition of proper scores.
\newblock \emph{Quarterly Journal of the Royal Meteorological Society},
  135\penalty0 (643):\penalty0 1512--1519, 2009.

\bibitem[Buja et~al.(2005)Buja, Stuetzle, and Shen]{Buja}
A.~Buja, W.~Stuetzle, and Y.~Shen.
\newblock Loss functions for binary class probability estimation and
  classification: Structure and applications.
\newblock \emph{(Technical Report) University of Pennsylvania}, 2005.

\bibitem[Burden and Faires(2010)]{book:NumericalAnalysis}
Richard~L. Burden and J.~Douglas Faires.
\newblock \emph{Numerical Analysis (Ninth Edition)}.
\newblock Brooks Cole, Boston, 2010.

\bibitem[Cover and Hart(1967)]{NNClassification}
T.~Cover and P.~Hart.
\newblock Nearest neighbor pattern classification.
\newblock \emph{IEEE Transactions on Information Theory}, 13\penalty0
  (1):\penalty0 21--27, 1967.

\bibitem[David~G(1969)]{book:OptVector}
Luenberger David~G.
\newblock \emph{Optimization By Vector Space Methods}.
\newblock John Wiley and Sons, New York, 1969.

\bibitem[Dawid(1982)]{Dawid1981}
A.P. Dawid.
\newblock The well-calibrated bayesian.
\newblock \emph{Journal of the American. Statistical Association}, 77:\penalty0
  605–--610, 1982.

\bibitem[DeGroot(1979)]{DeGroot2}
M.H. DeGroot.
\newblock Comments on lindley, et al.
\newblock \emph{Journal of Royal Statistical Society (A)}, 142:\penalty0
  172--173, 1979.

\bibitem[DeGroot and Fienberg(1982)]{DeGrootBook}
M.H. DeGroot and S.E. Fienberg.
\newblock Assessing probability assessors: calibration and refinement.
\newblock \emph{Statistical Decision Theory and Related Topics III},
  1:\penalty0 291--314, 1982.

\bibitem[DeGroot and Fienberg(1983)]{DeGroot}
Morris~H. DeGroot and Stephen~E. Fienberg.
\newblock The comparison and evaluation of forecasters.
\newblock \emph{The Statistician}, 32:\penalty0 14--22, 1983.

\bibitem[Fleuret and Guyon(2004)]{FastBinaryFeatureSelection}
Francois Fleuret and Isabelle Guyon.
\newblock Fast binary feature selection with conditional mutual information.
\newblock \emph{Journal of Machine Learning Research}, 5:\penalty0 1531--1555,
  2004.

\bibitem[Freund and Schapire(1997)]{freund}
Y.~Freund and R.~Schapire.
\newblock A decision-theoretic generalization of on-line learning and an
  application to boosting.
\newblock \emph{Journal of Computer and System Sciences}, 55:\penalty0 119–139,
  1997.

\bibitem[Friedman et~al.(2000)Friedman, Hastie, and Tibshirani]{friedman}
J.~Friedman, T.~Hastie, and R.~Tibshirani.
\newblock Additive logistic regression: A statistical view of boosting.
\newblock \emph{Annals of Statistics}, 28:\penalty0 337--407, 2000.

\bibitem[Fukunaga(1990)]{book:Fukunaga}
Keinosuke Fukunaga.
\newblock \emph{Introduction to Statistical Pattern Recognition, Second
  Edition}.
\newblock Academic Press, San Diego, 1990.

\bibitem[Gneiting and Raftery(2007)]{Raftery}
T.~Gneiting and A.E. Raftery.
\newblock Strictly proper scoring rules, prediction, and estimation.
\newblock \emph{Journal of the American Statistical Association}, 102:\penalty0
  359–--378, 2007.

\bibitem[Gneiting et~al.(2005)Gneiting, Raftery, Westveld, and
  Goldman]{Tilmann2005}
T.~Gneiting, A.~E. Raftery, A.~H. Westveld, and T.~Goldman.
\newblock Calibrated probabilistic forecasting using ensemble model output
  statistics and minimum crps estimation.
\newblock \emph{Monthly Weather Review}, 133:\penalty0 1098--1118, 2005.

\bibitem[Gneiting et~al.(2007)Gneiting, Balabdaoui, and Raftery]{Tilmann2007}
Tilmann Gneiting, Fadoua Balabdaoui, and Adrian~E. Raftery.
\newblock Probabilistic forecasts, calibration and sharpness.
\newblock \emph{Journal of the Royal Statistical Society Series B}, pages
  243--268, 2007.

\bibitem[Jaynes and Bretthorst(2003)]{book:Jaynes}
E.~T. Jaynes and G.~Larry Bretthorst.
\newblock \emph{Probability Theory: The Logic of Science}.
\newblock Cambridge University Press, Cambridge, 2003.

\bibitem[Lin(1991)]{JenShannonLin}
Jianhua Lin.
\newblock Divergence measures based on the shannon entropy.
\newblock \emph{IEEE Transactions on Information Theory}, 37:\penalty0
  145--151, 1991.

\bibitem[Masnadi-Shirazi and Vasconcelos(2008)]{HamedNunoLossDesign}
Hamed Masnadi-Shirazi and Nuno Vasconcelos.
\newblock On the design of loss functions for classification: theory,
  robustness to outliers, and savageboost.
\newblock In \emph{Advances in Neural Information Processing Systems}, pages
  1049--1056. MIT Press, 2008.

\bibitem[Masnadi-Shirazi et~al.(2010)Masnadi-Shirazi, Mahadevan, and
  Vasconcelos]{HamedNunoTangent}
Hamed Masnadi-Shirazi, Vijay Mahadevan, and Nuno Vasconcelos.
\newblock On the design of robust classifiers for computer vision.
\newblock In \emph{Computer Vision and Pattern Recognition, IEEE Conference
  on}, pages 779--786, 2010.

\bibitem[Murphy(1972)]{Murphy1972}
A.H. Murphy.
\newblock Scalar and vector partitions of the probability score: part i.
  two-state situation.
\newblock \emph{Journal of applied Meteorology}, 11:\penalty0 273--82, 1972.

\bibitem[Niculescu-Mizil and Caruana(2005)]{Caruana}
A.~Niculescu-Mizil and R.~Caruana.
\newblock Obtaining calibrated probabilities from boosting.
\newblock In \emph{Uncertainty in Artificial Intelligence}, 2005.

\bibitem[Peng et~al.(2005)Peng, Long, and Ding]{Pengfeatureselection}
Hanchuan Peng, Fuhui Long, and Chris Ding.
\newblock Feature selection based on mutual information: criteria of
  max-dependency, max-relevance, and min-redundancy.
\newblock \emph{IEEE Transactions on Pattern Analysis and Machine
  Intelligence}, 27:\penalty0 1226--1238, 2005.

\bibitem[Platt(2000)]{Platt}
J.~Platt.
\newblock Probabilistic outputs for support vector machines and comparison to
  regularized likelihood methods.
\newblock In \emph{Adv. in Large Margin Classifiers}, pages 61--74, 2000.

\bibitem[Reid and Williamson(2010)]{Reid}
Mark Reid and Robert Williamson.
\newblock Composite binary losses.
\newblock \emph{The Journal of Machine Learning Research}, 11:\penalty0
  2387--2422, 2010.

\bibitem[Sanders(1963)]{Sanders1963}
F.~Sanders.
\newblock On subjective probability forecasting.
\newblock \emph{Journal of applied Meteorology}, 2:\penalty0 191--201, 1963.

\bibitem[Savage(1971)]{Savage}
Leonard~J. Savage.
\newblock The elicitation of personal probabilities and expectations.
\newblock \emph{Journal of The American Statistical Association}, 66:\penalty0
  783--801, 1971.

\bibitem[Schervish(1989)]{Schervish1989}
M.J. Schervish.
\newblock A general method for comparing probability assessors.
\newblock \emph{Annals of Statistics}, 17:\penalty0 1856--1879, 1989.

\bibitem[Tang et~al.(2000)Tang, Yang, Liu, and Ma]{book:WaveletTheory}
Y.Y. Tang, L.H. Yang, J.~Liu, and H.~Ma.
\newblock \emph{Wavelet Theory and Its Application to Pattern Recognition}.
\newblock World Scientific Publishing, Singopore, 2000.

\bibitem[Vapnik(1998)]{book:vapnik}
Vladimir~N. Vapnik.
\newblock \emph{Statistical Learning Theory}.
\newblock John Wiley Sons Inc, 1998.

\bibitem[Vasconcelos and Vasconcelos(2009)]{NunoNaturalFeatures}
Manuela Vasconcelos and Nuno Vasconcelos.
\newblock Natural image statistics and low-complexity feature selection.
\newblock \emph{Pattern Analysis and Machine Intelligence, IEEE Transactions
  on}, 31:\penalty0 228--244, 2009.

\bibitem[Vasconcelos(2002)]{NUNOMaxDiversityNIPS}
Nuno Vasconcelos.
\newblock Feature selection by maximum marginal diversity.
\newblock In \emph{Advances in Neural Information Processing Systems}, pages
  1351--1358, 2002.

\bibitem[Vasconcelos(2003)]{NUNOMaxDiversityCVPR}
Nuno Vasconcelos.
\newblock Feature selection by maximum marginal diversity: optimality and
  implications for visual recognition.
\newblock In \emph{Computer Vision and Pattern Recognition, IEEE Computer
  Society Conference}, pages 762--769, 2003.

\bibitem[Wilks(2006)]{Wilks2006}
D.~S. Wilks.
\newblock Comparison of ensemble-mos methods in the lorenz '96 setting.
\newblock \emph{Meteorological Applications}, 13\penalty0 (3):\penalty0
  243--256, 2006.

\bibitem[Zhang(2004)]{Zhang}
Tong Zhang.
\newblock Statistical behavior and consistency of classification methods based
  on convex risk minimization.
\newblock \emph{Annals of Statistics}, 32:\penalty0 56--85, 2004.

\end{thebibliography}

\end{document}